\pdfoutput=1

\documentclass[11pt,a4paper]{article}
\usepackage[hyphens]{url}
\usepackage[hyperindex,breaklinks]{hyperref}
\usepackage[hyperref]{conll-2019}
\usepackage{times}
\usepackage{latexsym}
\usepackage{tabulary}
\usepackage{amsmath}
\newcommand{\argmin}{\mathop{\mathrm{argmin}}}

\aclfinalcopy % Uncomment this line for the final submission

\usepackage{tikz}
\usepackage{pgfplots}
\usepackage{xcolor}
\usepackage{color, colortbl}
\usepgfplotslibrary{groupplots}
\usepackage{url}
\usepackage{multirow}
\definecolor{bblue}{HTML}{4F81BD}
\definecolor{oorange}{HTML}{F4C842}
\definecolor{rred}{HTML}{C0504D}
\definecolor{ggreen}{HTML}{9BBB59}
\definecolor{ppurple}{HTML}{9F4C7C}
\definecolor{darkgreen}{HTML}{228B22}
\definecolor{cred}{HTML}{D81B60}
\definecolor{cblue}{HTML}{1E88E5}
\definecolor{cyellow}{HTML}{FFC107}
\definecolor{nred}{HTML}{e41a1c}
\definecolor{nblue}{HTML}{377eb8}
\definecolor{ngreen}{HTML}{4daf4a}

\newcommand{\draftonly}[1]{#1} 
\renewcommand{\draftonly}[1]{}

\usepackage{xspace}

\newcommand{\lang}[1]{\textsc{#1}}
\newcommand{\ara}{\lang{ara}\xspace}
\newcommand{\cmn}{\lang{cmn}\xspace}
\newcommand{\deu}{\lang{deu}\xspace}
\newcommand{\fra}{\lang{fra}\xspace}
\newcommand{\eng}{\lang{eng}\xspace}

\newcommand{\fin}{\lang{fin}\xspace}
\newcommand{\heb}{\lang{heb}\xspace}
\newcommand{\hrv}{\lang{hrv}\xspace}
\newcommand{\hun}{\lang{hun}\xspace}
\newcommand{\ita}{\lang{ita}\xspace}
\newcommand{\jpn}{\lang{jpn}\xspace}
\newcommand{\nld}{\lang{nld}\xspace}
\newcommand{\rus}{\lang{rus}\xspace}
\newcommand{\spa}{\lang{spa}\xspace}
\newcommand{\vie}{\lang{vie}\xspace}
\newcommand{\kaz}{\lang{kaz}\xspace}
\newcommand{\tur}{\lang{tur}\xspace}
\newcommand{\swe}{\lang{swe}\xspace}
\newcommand{\por}{\lang{por}\xspace}
\newcommand{\uig}{\lang{uig}\xspace}

\newcommand{\cwrs}{\textsc{cwr}s\xspace}
\title{Low-Resource Parsing with Crosslingual Contextualized Representations}

\author{
    Phoebe Mulcaire$^{\heartsuit}$\thanks{\quad Equal contribution. Random order.} \quad
	Jungo Kasai$^{\heartsuit}$\footnotemark[1] \quad Noah A. Smith$^{\heartsuit \diamondsuit}$\\
    $^{\heartsuit}$Paul G.~Allen School of Computer Science \& Engineering,\\University of Washington, Seattle, WA, USA \\
    $^{\diamondsuit}$Allen Institute for Artificial Intelligence, Seattle, WA, USA \\
    {\tt \{pmulc,jkasai,nasmith\}@cs.washington.edu}
}

\date{}

\begin{document}
\maketitle

\begin{abstract}
Despite advances in dependency parsing, languages with small treebanks still present challenges.
We assess recent approaches to multilingual contextual word representations (\cwrs), and compare them for crosslingual transfer from a language with a large treebank to a language with a small or nonexistent treebank, by sharing parameters between languages in the parser itself.
We experiment with a diverse selection of languages in both simulated and truly low-resource scenarios,  and show that multilingual \cwrs greatly facilitate low-resource dependency parsing even without crosslingual supervision such as dictionaries or parallel text.
Furthermore, we examine the non-contextual part of the learned language models (which we call a ``decontextual probe") to demonstrate that polyglot language models better encode crosslingual lexical correspondence compared to aligned monolingual language models.
This analysis provides further evidence that polyglot training is an effective approach to crosslingual transfer.
\end{abstract}

\section{Introduction}
Dependency parsing has achieved new states of the art using  distributed word representations in neural networks, trained with large amounts of annotated data \cite{dozatmanning2017,dozat-qi-manning:2017:K17-3,ma-EtAl:2018:Long2,Che2018ElmoUD}.
However, many languages are low-resource, with small or no treebanks, which presents a severe challenge in developing accurate parsing systems in those languages. 
One way to address this problem is with a crosslingual solution that makes use of a language with a large treebank and raw text in both languages.
The hypothesis behind this approach is that, although each language is unique, different languages manifest similar characteristics (e.g., morphological, lexical, syntactic) which can be exploited by training a single \textit{polyglot} model with data from multiple languages \cite{ammarthesis}.

Recent work has extended contextual word representations (\cwrs) multilingually either by training a polyglot language model (LM) on a mixture of data from multiple languages (\textit{joint training} approach; \citealp{mulcaire_NAACL2019}; \citealp{LampleConneau2019}) 
or by aligning multiple monolingual language models crosslingually (\textit{retrofitting} approach; \citealp{Schuster2019CrossLingual,aldarmaki_diab2019}).
These multilingual representations have been shown to facilitate crosslingual transfer on several tasks, including Universal Dependencies parsing and natural language inference.
In this work, we assess these two types of methods by using them for low-resource dependency parsing, and discover that the joint training approach substantially outperforms the retrofitting approach.
We further apply multilingual \cwrs produced by the joint training approach to diverse languages, and show that it is still effective in transfer between distant languages, though we find that phylogenetically related source languages are generally more helpful.

We hypothesize that joint polyglot training is more successful than retrofitting because it induces a degree of lexical correspondence between languages that the linear transformation used in retrofitting methods cannot capture. 
To test this hypothesis, we design a \textit{decontextual probe}. We \textit{decontextualize} \cwrs into non-contextual word vectors that retain much of \cwrs' task-performance benefit, and evaluate the crosslingual transferability of language models via word translation.
In our decontextualization framework, we use a single LSTM cell without recurrence to obtain a context-independent vector, thereby allowing for a direct probe into the LSTM networks independent of a particular corpus.
We show that decontextualized vectors from the joint training approach yield representations that score higher on a word translation task than the retrofitting approach or word type vectors such as fastText \cite{bojanowski2017enriching}.
This finding provides evidence that polyglot language models encode crosslingual similarity, specifically crosslingual lexical correspondence, that a linear alignment between monolingual language models does not. 

\section{Models}
\label{sec:models}
We examine crosslingual solutions to low-resource dependency parsing, which make crucial use of multilingual \cwrs.
All models are implemented in AllenNLP, version 0.7.2 \cite{Gardner2017AllenNLP} and the hyperparameters and training details are given in the appendix.

\subsection{Multilingual \cwrs}
\label{sec:multilingual_cwrs}
Prior methods to produce multilingual contextual word representations (\cwrs) can be categorized into two major classes, which we call \textit{joint training} and \textit{retrofitting}.\footnote{This term was originally used by \citet{faruqui2015retrofitting} to describe updates to word vectors, after estimating them from corpora, using semantic lexicons.  We generalize it to capture the notion of a separate update to fit something other than the original data, applied after conventional training.}
The joint training approach trains a single polgylot language model (LM) on a mixture of texts in multiple languages \cite{mulcaire_NAACL2019,LampleConneau2019,devlins2019bert},\footnote{Multilingual BERT is documented in \url{https://github.com/google-research/bert/blob/master/multilingual.md}.}
while the retrofitting approach trains separate LMs on each language and aligns the learned representations later \cite{Schuster2019CrossLingual,aldarmaki_diab2019}. 
We compare example approaches from these two classes using the same LM training data, and discover that the joint training approach generally yields better performance in low-resource dependency parsing, even without crosslingual supervision. 

\paragraph{Retrofitting Approach}
Following \citet{Schuster2019CrossLingual}, we first train a bidirectional LM with two-layer LSTMs on top of character CNNs for each language (ELMo, \citealp{Peters2018}), and then align the monolingual LMs across languages.
Denote the hidden state in the $j$th layer for word $i$ in context $c$ by $\mathbf{h}_{i,c}^{(j)}$.
We use a trainable weighted average of the three layers (character-CNN and two LSTM layers) to compute the contextual representation $\mathbf{e}_{i,c}$ for the word: $\mathbf{e}_{i,c} = \sum_{j=0}^2 \lambda_j \mathbf{h}_{i,c}^{(j)}$ \cite{Peters2018}.\footnote{\citet{Schuster2019CrossLingual} only used the first LSTM layer, but we found a performance benefit from using all layers in preliminary results.}
In the first step, we compute an ``anchor" $\mathbf{h}_{i}^{(j)}$ for each word by averaging $\mathbf{h}_{i,c}^{(j)}$ over all occurrences in an LM corpus.
We then apply a standard dictionary-based technique\footnote{\citet{conneau2017word} developed an unsupervised alignment technique that does not require a dictionary. We found that their unsupervised alignment yielded substantially degraded performance in downstream parsing in line with the findings of \citet{Schuster2019CrossLingual}.} to create multilingual word embeddings \cite{Mikolov2013ExploitingSA,conneau2017word}.
In particular, suppose that we have a word-translation dictionary from source language $s$ to target language $t$.
Let $\mathbf{H}_s^{(j)},\mathbf{H}_t^{(j)}$ be matrices whose columns are the anchors in the $j$th layer for the source and corresponding target words in the dictionary.
For each layer $j$, find the linear transformation $\mathbf{W}^{*(j)}$ such that
\begin{align*}
\mathbf{W}^{* (j)} = \argmin_{\mathbf{W}} || \mathbf{W} \mathbf{H}_s ^{(j)}  -\mathbf{H}_{t}^{(j)} ||_{F}
\end{align*}
The linear transformations are then used to map the LM hidden states for the source language to the target LM space. 
Specifically, contextual representations for the source and target languages are computed by 
$\sum_{j=0}^2 \lambda_j \mathbf{W}^{*(j)} \mathbf{h}_{i,c}^{(j)}$ and $\sum_{j=0}^2 \lambda_j\mathbf{h}_{i,c}^{(j)}$ respectively.
We use publicly available dictionaries from \citet{conneau2017word}\footnote{\url{https://github.com/facebookresearch/MUSE\#ground-truth-bilingual-dictionaries}} and align all languages to the English LM space, again following \citet{Schuster2019CrossLingual}.

\paragraph{Joint Training Approach}
Another approach to multilingual \cwrs is to train a single LM on multiple languages \cite{Tsvetkov2016, Ragni2016MultiLanguageNN, ostling-tiedemann-2017-continuous}.
We train a single bidirectional LM with charater CNNs and two-layer LSTMs on multiple languages (Rosita, \citealp{mulcaire_NAACL2019}).
We then use the polyglot LM to provide contextual representations.
Similarly to the retrofitting approach, we represent word $i$ in context $c$ as a trainable weighted average of the hidden states in the trained polyglot LM: $\sum_{j=0}^2 \lambda_j \mathbf{h}_{i,c}^{(j)}$.
In contrast to retrofitting, crosslinguality is learned implicitly by sharing all network parameters during LM training; no crosslingual dictionaries are used.

\paragraph{Refinement after Joint Training}
It is possible to combine the two approaches above; the alignment procedure used in the retrofitting approach can serve as a refinement step on top of an already-polyglot language model.
We will see only a limited gain in parsing performance from this refinement in our experiments, suggesting that polyglot LMs are already producing high-quality multilingual \cwrs even without crosslingual dictionary supervision.

\paragraph{FastText Baseline} We also compare the multilingual \cwrs to a subword-based, non-contextual word embedding baseline. We train 300-dimensional word vectors on the same LM data using the fastText method \cite{bojanowski2017enriching}, and use the same bilingual dictionaries to align them \cite{conneau2017word}. 
\subsection{Dependency Parsers}
We train polyglot parsers for multiple languages \cite{Ammar2016ManyLO}
on top of multilingual \cwrs. 
All parser parameters are shared between the source and target languages.
\citet{Ammar2016ManyLO} suggest that sharing parameters between languages can alleviate the low-resource problem in syntactic parsing, but their experiments are limited to (relatively similar) European languages.
\citet{mulcaire_NAACL2019} also include experiments with dependency parsing using polyglot contextual representations between two language pairs (English/Chinese and English/Arabic), but focus on high-resource tasks.
Here we explore a wider range of languages, and analyze the particular efficacy of a crosslingual approach to dependency parsing in a low-resource setting.

We use a strong graph-based dependency parser with BiLSTM and biaffine attention \cite{dozatmanning2017}, which is also used in related work \cite{Schuster2019CrossLingual, mulcaire_NAACL2019}.
Crucially, our parser only takes as input word representations.
Universal parts of speech have been shown useful for low-resource dependency parsing \cite{duong-EtAl:2015:ACL-IJCNLP,Ammar2016ManyLO,Ahmad2018NearOF}, but many realistic low-resource scenarios lack reliable part-of-speech taggers; here, we do not use parts of speech as input, and thus avoid the error-prone part-of-speech tagging pipeline.
For the fastText baseline, word embeddings are not updated during training, to preserve crosslingual alignment \cite{Ammar2016ManyLO}.

{
\renewcommand{\tabcolsep}{4.65pt}
\begin{table}[h]
    \small
    \begin{flushleft}
    \begin{tabulary}{0.5\textwidth}{l|lll|r}
    \hline
    Lang & Code & Genus & WALS 81A & Size \\ \hline\hline
    English & \eng & Germanic & SVO & -- \\ \hline
    Arabic & \ara & Semitic & VSO/SVO &\\
    Hebrew & \heb & Semitic & SVO & \multirow{-2}{*}{5241}\\
%\rowcolor[gray]{0.9} Estonian & est & Finnic & SVO\\
%\rowcolor[gray]{0.9} Finnish & fin & Finnic &SVO \\
    \rowcolor[gray]{0.9} Croatian & \hrv & Slavic & SVO &\\
    \rowcolor[gray]{0.9} Russian  & \rus & Slavic & SVO & \multirow{-2}{*}{6983}\\
    Dutch   & \nld & Germanic & SOV/SVO & \\
    German  & \deu & Germanic & SOV/SVO & \multirow{-2}{*}{12269}\\
    \rowcolor[gray]{0.9} Spanish & \spa & Romance & SVO & \\
    \rowcolor[gray]{0.9} Italian & \ita & Romance & SVO & \multirow{-2}{*}{12543}\\
    Chinese  & \cmn & Chinese  & SVO & \\
    Japanese & \jpn & Japanese & SOV & \multirow{-2}{*}{3997}\\ \hline
    %\multicolumn{5}{c}{Truly Low Resource Pairs} \\ \hline
    \textbf{Hungarian}  & \hun & Ugric   & SOV/SVO   &  910\\
    Finnish     & \fin & Finnic  & SVO       & 12217\\ 
    \rowcolor[gray]{0.9} \textbf{Vietnamese}    & \vie & Viet-Muong  & SVO & 1400\\
    \textbf{Uyghur}    & \uig & Turkic & SOV  & 1656 \\
    \textbf{Kazakh}    & \kaz & Turkic  & SOV & 31\\
    Turkish    & \tur & Turkic  & SOV & 3685\\
    \end{tabulary}
    \vspace{-1mm}
    \caption{List of the languages used in our UD v2.2 experiments. Each shaded/unshaded section corresponds to a pair of ``related'' languages. WALS 81A denotes Feature 81A in WALS, Order of Subject, Object, and Verb \cite{wals}. ``Size'' represents the downsampled size in \# of sentences used for source treebanks. The four languages in bold face are truly low resource languages ($<2000$ trees).}
    \label{tab:lang_list}
    \end{flushleft}
\end{table}
}

\section{Experiments}
We first conduct a set of experiments to assess the efficacy of multilingual \cwrs for low-resource dependency parsing.
\subsection{Zero-Target Dependency Parsing}
Following prior work on low-resource dependency parsing and crosslingual transfer \cite{zhang-barzilay-2015-hierarchical,guo-etal-2015-cross,Ammar2016ManyLO,Schuster2019CrossLingual}, we conduct multi-source experiments on six languages (German, Spanish, French, Italian, Portuguese, and Swedish) from Google universal dependency treebank version 2.0 \cite{mcdonald-etal-2013-universal}.\footnote{\url{http://github.com/ryanmcd/uni-dep-tb}}
We train language models on the six languages and English to produce multilingual \cwrs.
For each tested language, we train a polyglot parser with the multilingual \cwrs on the five other languages and English, and apply the parser to the test data for the target language.
Importantly, the parsing annotation scheme is shared among the seven languages.
Our results will show that the joint training approach for \cwrs %(Rosita)
substantially outperforms the retrofitting approach. % (alignment of ELMos).

\subsection{Diverse Low-Resource Parsing}
\label{sec:lang_pairs}
The previous experiment compares the joint training and retrofitting approaches in low-resource dependency parsing only for relatively similar languages.
In order to study the effectiveness more extensively, we apply it to a more typologically diverse set of languages.
We use five pairs of languages for ``low-resource simulations,'' in which we reduce the size of a large treebank, and four languages for ``true low-resource experiments,'' where only small UD treebanks are available, allowing us to compare to other work in the low-resource condition (Table \ref{tab:lang_list}).
Following \citet{delhoneux-EtAl:2018:EMNLP}, we selected these language pairs to represent linguistic diversity.
For each target language,  we produce multilingual \cwrs by training a polyglot language model with its related language (e.g., Arabic and Hebrew) as well as English (e.g., Arabic and English).
We then train a polyglot dependency parser on each language pair and assess the crosslingual transfer in terms of target parsing accuracy. 

Each pair of related languages shares features like word order, morphology, or script. For example, Arabic and Hebrew are similar in their rich transfixing morphology \cite{delhoneux-EtAl:2018:EMNLP}, and Dutch and German share most of their word order features.
We chose Chinese and Japanese as an example of a language pair which does \textit{not} share a language family but does share characters.

We chose Hungarian, Vietnamese, Uyghur, and Kazakh as true low-resource target languages because they had comparatively small amounts of annotated text in the UD corpus (Vietnamese: 1,400 sentences, 20,285 tokens; Hungarian: 910 sentences, 20,166 tokens; Uyghur: 1,656 sentences, 19,262 tokens; Kazakh: 31 sentences, 529 tokens;), yet had convenient sources of  text for LM pretraining \cite{zeman-EtAl:2018:K18-2}.\footnote{The one exception is Uyghur where we only have 3M words in the raw LM data from \citet{zeman-EtAl:2018:K18-2}.}
Other small treebanks exist, but in most cases another larger treebank exists for the same language, making domain adaptation a more likely option than crosslingual transfer.
Also, recent work \cite{Che2018ElmoUD} using contextual embeddings was top-ranked for most of these languages in the CoNLL 2018 shared task on UD parsing \cite{zeman-EtAl:2018:K18-2}.\footnote{In Kazakh, \citet{Che2018ElmoUD} did not use \cwrs due to the extremely small treebank size.}

We use the same Universal Dependencies (UD) treebanks \cite{ud-2.2} and train/development/test splits as the CoNLL 2018 shared task \cite{zeman-EtAl:2018:K18-2}.\footnote{See Appendix for a list of UD treebanks used.}
The annotation scheme is again shared across languages, which facilitates crosslingual transfer. 
For each triple of two related languages and English, we downsample training and development data to match the language with the smallest treebank size.
This allows for fairer comparisons because within each triple, the source language for any parser will have the same amount of training data.
We further downsample sentences from the target train/development data to simulate low-resource scenarios. The ratio of training and development data is kept 5:1 throughout the simulations, and we denote the number of sentences in training data by $|D_{\tau}|$.
For testing, we use the CoNLL 2018 script on the gold word segmentations. 
For the truly low-resource languages, we also present results with word segmentations from the system outputs of \citet{Che2018ElmoUD} (\hun, \vie, \uig) and \citet{smith-etal-2018-82} (\kaz) for a direct comparison to those languages' best previously reported parsers.\footnote{System outputs for all shared task systems are available at
\url{https://lindat.mff.cuni.cz/repository/xmlui/handle/11234/1-2885}}

\section{Results and Discussion}
In this section we describe the results of the various parsing experiments.

\subsection{Zero-Target Parsing}
\begin{table*}
\small
\centering
\begin{tabular}{ l | ccccccc }
\hline
Model &  \deu & \spa & \fra & \ita & \por & \swe & AVG\\\hline
\citet{Schuster2019CrossLingual} (retrofitting) & 61.4 & 77.5 & \textbf{77.0} & 77.6 & 73.9 & 71.0 & 73.1\\ 
\citet{Schuster2019CrossLingual} (retrofitting, no dictionaries) & \textbf{61.7} & 76.6 & 76.3 & \textbf{77.1} & 69.1 & 54.2 & 69.2\\ \hline
fastText + Alignment &45.2 &68.5 & 62.8 & 58.9 & 61.1 & 50.4 & 57.8\\
ELMos + Alignment (retrofitting)&57.3 & 75.4 &73.7& 71.6 & 75.1 & 74.2&71.2\\
Rosita (joint training, no dictionaries) & 58.0 & \textbf{81.8}& 75.6 & 74.8 & \textbf{77.1} & 76.2&73.9 \\
Rosita + Refinement (joint training + retrofitting) & \textbf{61.7} & 79.7& 75.8 & 76.0& 76.8 &\textbf{76.7} & \textbf{74.5}\\
\end{tabular}
\caption{Zero-target results in LAS. Results reported in prior work (above the line) use an unknown amount of LM training data; all models below the line are limited to approximately 50M words per language.}
\label{tab:malopa}
\end{table*}

\begin{table*}
\small
\centering
\begin{tabular}{ l | ccccccc }
\hline
Model &  \deu & \spa & \fra & \ita & \por & \swe & AVG\\\hline
\citet{zhang-barzilay-2015-hierarchical} & 54.1& 68.3 &68.8& 69.4& 72.5 &62.5& 65.9\\
\citet{Guo2016ARL} &55.9&73.1 &71.0&71.2&78.6&69.5&69.9\\
\citet{Ammar2016ManyLO}  & 57.1 & 74.6 & 73.9 & 72.5 & 77.0 & 68.1 & 70.5\\
\citet{Schuster2019CrossLingual} (retrofitting) & \textbf{65.2} &  80.0 & \textbf{80.8} & \textbf{79.8} & 82.7 & 75.4 & 77.3\\ 
\citet{Schuster2019CrossLingual} (retrofitting, no dictionaries) & 64.1& 77.8  & 79.8 &  79.7&79.1 & 69.6 & 75.0\\ \hline
Rosita (joint training, no dictionaries) & 63.6 & \textbf{83.4} & 78.9& 77.8 & 83.0 &\textbf{79.6} &77.7\\
Rosita + Refinement (joint training + retrofitting)& 64.8 & 82.1 & 78.7 & 78.8 & \textbf{84.1} & 79.1 & \textbf{77.9}\\
\end{tabular}
\caption{Zero-target results in LAS with gold UPOS. }
\label{tab:malopa_gold}
\end{table*}

Table \ref{tab:malopa} shows results on zero-target dependency parsing.
First, we see that all \cwrs greatly improve upon the fastText baseline. 
The joint training approach (Rosita), which uses no dictionaries, consistently outperforms the dictionary-dependent retrofitting approach (ELMos+Alignment).
As discussed in the previous section, we can apply the alignment method to refine the already-polyglot Rosita using dictionaries.
However, we  observe a relatively limited gain in overall performance (74.5 vs.~73.9 LAS points), suggesting that Rosita (polyglot language model) is already developing useful multilingual \cwrs for parsing without crosslingual supervision.
Note that the degraded overall performance of our ELMo+Alignment compared to \citet{Schuster2019CrossLingual}'s reported results (71.2 vs.\ 73.1) is likely due to the significantly reduced amount of LM data we used in all of our experiments (50M words per language, an order of magnitude reduction from the full Wikipedia dumps used in \citet{Schuster2019CrossLingual}).
\citet{Schuster2019CrossLingual} (no dictionaries) is the same retrofitting approach as ELMos+Alignment except that the transformation matrices are learned in an unsupervised fashion without dictionaries \cite{conneau2017word}.
The absence of a dictionary yields much worse performance (69.2 vs.\ 73.1) in contrast with the joint training approach of Rosita, which also does not use a dictionary (73.9).

We also present results using gold universal part of speech to compare to previous work in Table \ref{tab:malopa_gold}.
We again see Rosita's effectiveness and a marginal benefit from refinement with dictionaries. It should also be noted that the reported results for French, Italian and German in  \citet{Schuster2019CrossLingual} outperform all results from our controlled comparison; this may be due to the use of abundant LM training data. Nevertheless, joint training, with or without refinement, performs best on average in both gold and predicted POS settings.

\subsection{Diverse Low-Resource Parsing}
\label{sec:sim_results}
\paragraph{Low-Resource Simulations}
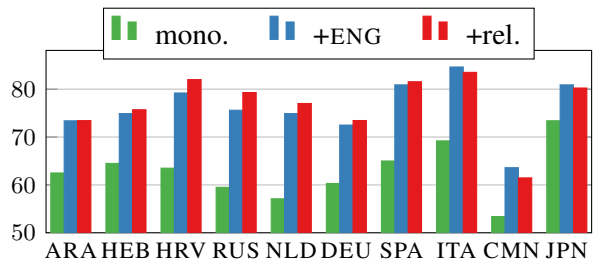
\begin{figure}
\pgfplotsset{compat=1.3}
\pgfplotsset{every y tick label/.append style={font=\small}}
\begin{flushleft}
\begin{tikzpicture}%[font=\large]
    \clip (-0.45,-0.41) rectangle (0.46\textwidth,3.1cm);
    \begin{axis}[
        width  = 0.5475\textwidth,
        height = 4cm,
        major x tick style = transparent,
        ybar=0pt,
        bar width=5pt,
        ymajorgrids = true,
        symbolic x coords={ara,heb,hrv,rus,nld,deu,spa,ita,cmn,jpn},
        xticklabels = {
			\strut \textsc{ara},	\strut \textsc{heb},\strut \textsc{hrv},\strut \textsc{rus},\strut \textsc{nld},\strut \textsc{deu},\strut \textsc{spa},\strut \textsc{ita},\strut \textsc{cmn},\strut \textsc{jpn}
        },
        xticklabel style={yshift=8pt},
        xtick = data,
        scaled y ticks = false,
        enlarge x limits=0.05,
        ymin=50,
        legend cell align=left,
        legend style={legend columns=3,
                at={(0.5,0.95)},
                anchor=south,
                column sep=1ex
        }
    ]
        \addplot[style={ngreen,fill=ngreen,mark=none}]
            coordinates {(ara,62.5)(heb,64.5)(hrv,63.5)(rus,59.5)(nld,57.1)(deu,60.3)(spa,65.0)(ita,69.2)(cmn,53.4)(jpn,73.4)};
        \addplot[style={nblue,fill=nblue,mark=none}]
            coordinates {(ara,73.39)(heb,74.9)(hrv,79.2)(rus,75.6)(nld,74.9)(deu,72.5)(spa,80.9)(ita,84.63)(cmn,63.6)(jpn,80.9)};
        \addplot[style={nred,fill=nred,mark=none}]
            coordinates {(ara,73.43)(heb,75.7)(hrv,82.0)(rus,79.3)(nld,77.01)(deu,73.45)(spa,81.55)(ita,83.51)(cmn,61.47)(jpn,80.24)};
        \legend{\strut{mono.~~~~},\strut{+\textsc{eng}~~~~},\strut{+rel.}}
    \end{axis};
\end{tikzpicture}
\vspace{-5mm}
\end{flushleft}
\caption{LAS for UD parsing results in a simulated low-resource setting where the size of the target language treebank ($|D_{\tau}|$) is set to 100 sentences.}
%\vspace{-0.1cm}
\label{fig:bar_chart}
\end{figure}

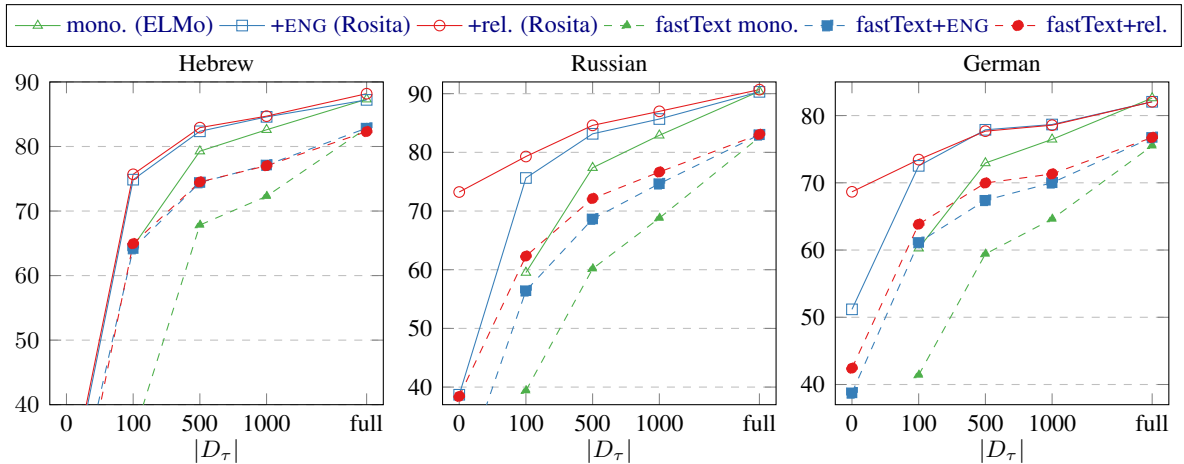
\begin{figure*}
\begin{tikzpicture}[scale=1.0]
\centering
\pgfplotsset{/pgf/number format/fixed, every tick label/.append style={font=\footnotesize}}
  \begin{groupplot}
    [
        anchor=north west,
        group style=
        {
            group size=3 by 1,
            horizontal sep=4.5ex,
            x descriptions at=edge bottom,
            y descriptions at=edge left,
        },
        height=\textwidth/2.73,
        width=\textwidth/2.68
    ]
    \nextgroupplot
        [
            title={\footnotesize Hebrew},
		    title style={yshift=-0.2cm},
            xmin=0, xmax=100,
            ymin=40, ymax=90,
            xtick={5, 25, 45, 65, 95},
            xticklabels={0, 100, 500, 1000, full},
            ytick={40, 50, 60, 70, 80, 90},
            legend to name=grouplegend,
			legend style={font=\small},
            legend pos=south east,
            legend style={font=\footnotesize},
            legend columns=6
			ylabel=LAS,
			xlabel={\small $|D_{\tau}|$},
		    ylabel style={yshift=-0.4cm}, %shifting the y line text
		    xlabel style={yshift=0.2cm},
            ymajorgrids=true,
            grid style=dashed,
		]
\addplot[color=ngreen, mark=triangle, mark size=2] 
coordinates {(25,64.53)(45,79.27)(65,82.59)(95,87.32)};
    \addlegendentry{mono. (ELMo)}
\addplot[color=nblue, mark=square, mark size=2]
    coordinates {(5, 23.76)(25, 74.86)(45, 82.35)(65, 84.59)(95, 87.22)};
    \addlegendentry{+\eng (Rosita)}
\addplot[color=nred, mark=o, mark size=2]
    coordinates {(5, 24.89)(25, 75.69)(45, 82.92)(65, 84.7)(95, 88.18)};
    \addlegendentry{+rel. (Rosita)}
\addplot[color=ngreen, mark=triangle*, dashed, mark size=2]
    coordinates { (25, 35.02)(45,67.83)(65,72.29)(95, 82.84)};
    \addlegendentry{fastText mono.}
\addplot[color=nblue, mark=square*, dashed,mark size=2]
    coordinates {(5, 18.42)(25, 64.19)(45,74.42)(65,77.13)(95, 82.84)};
    \addlegendentry{fastText+\eng}
\addplot[color=nred, mark=otimes*,dashed,mark size=2]
    coordinates {(5, 14.23)(25, 64.95)(45,74.51)(65,77.02)(95, 82.31)};
    \addlegendentry{fastText+rel.}
    
    \nextgroupplot
        [
            title={\footnotesize Russian},
		    title style={yshift=-0.2cm},
            xmin=0, xmax=100,
            ymin=37, ymax=92,
            xtick={5, 25, 45, 65, 95},
            xticklabels={0, 100, 500, 1000, full},
            ytick={30, 40, 50, 60, 70, 80, 90},
            yticklabels={30, 40, 50, 60, 70, 80, 90},
            legend style={font=\footnotesize},
            legend columns=6,
            ymajorgrids=true,
            grid style=dashed,
			xlabel={\small $|D_{\tau}|$},
		    xlabel style={yshift=0.2cm},
		]
\addplot[color=ngreen, mark=triangle, mark size=2] 
coordinates {(25,59.51)(45,77.38)(65,82.9)(95,90.4)};
\addplot[color=nblue, mark=square, mark size=2]
    coordinates {(5, 38.69)(25, 75.63)(45, 83.16)(65, 85.68)(95, 90.31)};
\addplot[color=nred, mark=o,mark size=2]
    coordinates {(5, 73.24)(25, 79.29)(45, 84.6)(65, 86.99)(95, 90.69)};
    
\addplot[color=ngreen, mark=triangle*, dashed,mark size=2] 
coordinates {(25,39.39)(45,60.17)(65,68.79)(95,82.58)};
\addplot[color=nblue, mark=square*, dashed,mark size=2]
    coordinates {(5, 23.44)(25, 56.42)(45, 68.65)(65, 74.66)(95, 83)};
\addplot[color=nred, mark=otimes*,dashed,mark size=2]
    coordinates {(5, 38.42)(25, 62.32)(45, 72.14)(65, 76.65)(95, 83.06)};
    \nextgroupplot
        [
            title={\footnotesize German},
		    title style={yshift=-0.2cm},
            xmin=0, xmax=100,
            ymin=37, ymax=85,
            xtick={5, 25, 45, 65, 95},
            xticklabels={0, 100, 500, 1000, full},
            ytick={30, 40, 50, 60, 70, 80},
            yticklabels={30, 40, 50, 60, 70, 80},
            legend pos=south east,
            legend style={font=\footnotesize},
            ymajorgrids=true,
            grid style=dashed,
			xlabel={\small $|D_{\tau}|$},
		    xlabel style={yshift=0.2cm},
		]
\addplot[color=ngreen, mark=triangle, mark size=2] 
coordinates {(25,60.26)(45,72.94)(65,76.46)(95,82.6)};
\addplot[color=nblue, mark=square, mark size=2]
    coordinates {(5, 51.18)(25, 72.52)(45, 77.88)(65, 78.67)(95, 82.05)};
\addplot[color=nred, mark=o,mark size=2]
    coordinates {(5, 68.66)(25, 73.45)(45, 77.68)(65, 78.57)(95, 82.05)};
\addplot[color=ngreen, mark=triangle*, dashed,mark size=2] 
coordinates {(25,41.41)(45,59.41)(65,64.61)(95,75.5)};
\addplot[color=nblue, mark=square*, dashed,mark size=2]
    coordinates {(5, 38.76)(25, 61.13)(45, 67.4)(65, 70.01)(95, 76.75)};
\addplot[color=nred, mark=otimes*,dashed,mark size=2]
    coordinates {(5, 42.43)(25, 63.84)(45, 69.97)(65, 71.32)(95, 76.74)};
  \end{groupplot}
  \node at (group c1r1.north east) [inner sep=0pt,anchor=north, yshift=6ex, xshift=16ex] {\ref{grouplegend}};
\end{tikzpicture}
\vspace{-3mm}
\caption{Plots of parsing performance vs.~target language treebank size for several example languages. The size 0 target treebank point indicates a parser trained \emph{only} on the source language treebank but with polyglot representations, allowing transfer to the target test treebank using no target language training trees. See Appendix for results with zero-target-treebank and intermediate size data ($|D_{\tau}| \in \{0,100,500,1000\}$) for all languages.
\label{fig:group_plot}}
\end{figure*}

Figure \ref{fig:bar_chart} shows simulated low-resource results.\footnote{A table with full details including different size simulations is provided in the appendix.}
Of greatest interest are the significant improvements over monolingual parsers when adding English or related-language data.
This improvement is consistent across languages and suggests that crosslingual transfer is a viable solution for a wide range of languages, even when (as in our case) language-specific tuning or annotated resources like parallel corpora or bilingual dictionaries are not available. 
See Figure \ref{fig:group_plot} for a visualization of the differences in performance with varying training size.
The polyglot advantage is minor when the target language treebank is large, but dramatic in the condition where the target language has only 100 sentences. 
The fastText approaches consistently underperform the language model approaches, but show the same pattern.

In addition, related-language polyglot (``+rel.'') outperforms English polyglot in most cases in the low-resource condition. The exceptions to this pattern are Italian (whose treebank is of a different genre from the Spanish one), and Japanese and Chinese, which differ significantly in morphology and word order. The \cmn/\jpn result suggests that such typological features influence the degree of crosslingual transfer more than orthographic properties like shared characters.
This result in crosslingual transfer also mirrors the observation from prior work \cite{gerz2018relation} that typological features of the language are predictive of \textit{monolingual} LM performance.
The related-language improvement also vanishes in the full-data condition (Figure \ref{fig:group_plot}), implying that the importance of shared linguistic features can be overcome with sufficient annotated data.
It is also noteworthy that variations in word order, such as the order of adjective and noun, do not affect performance: Italian, Arabic, and others use a noun-adjective order while English uses an adjective-noun order, but their +\eng and +rel. results are comparable.

The Croatian and Russian results are notable because of shared heritage but different scripts. Though Croatian uses the Latin alphabet and Russian uses Cyrillic, transfer between \textsc{hrv}$+$\rus is clearly more effective than \textsc{hrv}$+$\eng (82.00 vs.\ 79.21 LAS points when $|D_{\tau}|=100$).
This suggests that character-based LMs can implicitly learn to transliterate between related languages with different scripts, even without parallel supervision.

\paragraph{Truly Low Resource Languages}
\begin{table}
    \vspace{0mm}
    \small
    \begin{center}
    \begin{tabulary}{\textwidth}{l|cc}
    \hline
    Model &  gold & pred.\\\hline
    \multicolumn{3}{l}{Hungarian (\hun)}\\
    \citet{Che2018ElmoUD} (\hun, ensemble) & -- & 82.66\\
    \citet{Che2018ElmoUD} (\hun) & -- & 80.96\\
    ELMo (\hun) & 81.89 & 81.54\\
    Rosita (\hun+\eng) &  85.34& 84.89\\
    Rosita (\hun+\fin) &  \textbf{85.40}& \textbf{84.96}\\\hline
    \multicolumn{3}{l}{Vietnamese (\vie)}\\
    \citet{Che2018ElmoUD} (\vie, ensemble) & --& 55.22\\
    ELMo (\vie) &  62.67 & 55.72 \\
    Rosita (\vie+\eng) &  \textbf{63.07} & \textbf{56.42} \\ \hline
    \multicolumn{3}{l}{Uyghur (\uig)}\\
    \citet{Che2018ElmoUD} (\uig, ensemble) & --& \textbf{67.05}\\
    \citet{Che2018ElmoUD} (\uig) & --& 66.20 \\
    ELMo (\uig) &  66.64 & 63.98 \\
    Rosita (\uig+\eng) &  67.85 & 65.55 \\ 
    Rosita (\uig+\tur) &  \textbf{68.08} & 65.73 \\ \hline
    \citet{rosa-marecek-2018-cuni} (\kaz+\tur)& -- & 26.31\\
    \citet{smith-etal-2018-82} (\kaz+\tur) & --& 31.93 \\
    \citet{Schuster2019CrossLingual} (\kaz+\tur) &--& 36.98\\
    Rosita (\kaz+\eng) &  48.02& 46.03\\
    Rosita (\kaz+\tur) &  \textbf{53.98}& \textbf{51.96}\\
   \end{tabulary}
    \caption{LAS ($F_1$) comparison for truly low-resource languages. The gold and pred.\ columns show results under gold segmentation and predicted segmentation. 
    The languages in the parentheses indicate the languages used in parser training.
    }
    \label{tab:truly_low_resource}
    \end{center}
\end{table}
Finally we present ``true low-resource'' experiments for four languages in which little UD data is available (see Section \ref{sec:lang_pairs}). Table \ref{tab:truly_low_resource} shows these results.
Consistent with our simulations, our parsers on top of Rosita (multilingual \cwrs from the joint training approach) substantially outperform the parsers with ELMos (monolingual \cwrs) in all languages, and establish a new state of the art in Hungarian, Vietnamese, and Kazakh.
Consistent with our simulations, we see that training parsers with the target's related language is more effective than with the more distant language, English.
It is particularly noteworthy that the Rosita models, which do not use a parallel corpus or dictionary, dramatically improve over the best previously reported result from \citet{Schuster2019CrossLingual} when either the related language of Turkish (51.96 vs.\ 36.98) or even the more distant language of English (46.03 v.s.\ 36.98) is used.
\citet{Schuster2019CrossLingual} aligned the monolingual ELMos for Kazakh and Turkish using the \kaz-\tur dictionary that \citet{rosa-marecek-2018-cuni} derived from parallel text. 
This result further corroborates our finding that the joint training approach to multilingual \cwrs is more effective than retrofitting monolingual LMs. 

\subsection{Comparison to Multilingual BERT Embeddings}

We also evaluate the diverse low-resource language pairs using pretrained multilingual BERT \cite{devlins2019bert} as text embeddings (Figure \ref{fig:bert_bar_chart}).
Here, the same language model (multilingual cased BERT,\footnote{Available at \url{https://github.com/google-research/bert/}} covering 104 languages) is used for all parsers, with the only variation being in the training treebanks provided to each parser. Parsers are trained using the same hyperparameters and data as in Section \ref{sec:lang_pairs}.\footnote{AllenNLP version 0.9.0 was used for these experiments.}

There are two critical differences  from our previous experiments: multilingual  BERT is trained on much larger amounts of Wikipedia data compared to other LMs used in this work, and the WordPiece vocabulary \cite{wu2016google} used in the cased multilingual BERT model has been shown to have a distribution skewed toward Latin alphabets \cite{acs_2019}.
These results are thus not directly comparable to those in Figure \ref{fig:bar_chart}; nevertheless, it is interesting to see that the results obtained with ELMo-like LMs are comparable to and in some cases better than results using a BERT model trained on over a hundred languages.
Our results broadly fit with those of \citet{pires2019multilingua}, who found that multilingual BERT was useful for zero-shot crosslingual syntactic transfer. In particular, we find nearly no performance benefit from cross-script transfer using BERT in a language pair (English-Japanese) for which they reported poor performance in zero-shot transfer, contrary to our results using Rosita (Section \ref{sec:sim_results}).

\begin{figure}
\pgfplotsset{compat=1.3}
\pgfplotsset{every y tick label/.append style={font=\small}}
\begin{flushleft}
\begin{tikzpicture}%[font=\large]
    \clip (-0.45,-0.41) rectangle (0.46\textwidth,3.1cm);
    \begin{axis}[
        width  = 0.5475\textwidth,
        height = 4cm,
        major x tick style = transparent,
        ybar=0pt,
        bar width=5pt,
        ymajorgrids = true,
        symbolic x coords={ara,heb,hrv,rus,nld,deu,spa,ita,cmn,jpn},
        xticklabels = {
			\strut \textsc{ara},	\strut \textsc{heb},\strut \textsc{hrv},\strut \textsc{rus},\strut \textsc{nld},\strut \textsc{deu},\strut \textsc{spa},\strut \textsc{ita},\strut \textsc{cmn},\strut \textsc{jpn}
        },
        xticklabel style={yshift=8pt},
        xtick = data,
        scaled y ticks = false,
        enlarge x limits=0.05,
        ymin=45,
        legend cell align=left,
        legend style={legend columns=3,
                at={(0.5,0.95)},
                anchor=south,
                column sep=1ex
        }
    ]
        \addplot[style={ngreen,fill=ngreen,mark=none}]
            coordinates {(ara,63.35)(heb,64.11)(hrv,61.91)(rus,56.32)(nld,57.96)(deu,60.58)(spa,67.39)(ita,70.35)(cmn,49.95)(jpn,71.80)};
        \addplot[style={nblue,fill=nblue,mark=none}]
            coordinates {(ara,81.35)(heb,71.97)(hrv,74.49)(rus,69.74)(nld,72.80)(deu,72.80)(spa,77.88)(ita,83.31)(cmn,59.62)(jpn,72.68)};
        \addplot[style={nred,fill=nred,mark=none}]
            coordinates {(ara,69.28)(heb,72.10)(hrv,79.60)(rus,76.32)(nld,75.53)(deu,71.76)(spa,79.63)(ita,81.83)(cmn,52.71)(jpn,71.41)};
        \legend{\strut{mono.~~~~},\strut{+\textsc{eng}~~~~},\strut{+rel.}}
    \end{axis};
\end{tikzpicture}
\vspace{-6mm}
\end{flushleft}
\caption{LAS for UD parsing results in a simulated low-resource setting (($|D_{\tau}|=100$) using multilingual BERT embeddings in place of Rosita. Cf. Figure \ref{fig:bar_chart}.}
%\vspace{-0.1cm}
\label{fig:bert_bar_chart}
\end{figure}

\section{Decontextual Probe}
We saw the success of the joint polyglot training for multilingual \cwrs over the retrofitting approach in the previous section.
We hypothesize that \cwrs from joint training provide useful representations for parsers by inducing \textit{nonlinear} similarity in the vector spaces of different languages that we cannot retrieve with a simple alignment of monolingual pretrained language models.
In order to test this hypothesis, we conduct a \textit{decontextual probe} comprised of two steps.
The decontextualization step effectively distills \cwrs into word type vectors, where each unique word is mapped to exactly one embedding regardless of the context. 
We then conduct linear transformation-based word translation \cite{Mikolov2013ExploitingSA} on the decontextualized vectors to quantify the degree of crosslingual similarity in the multilingual \cwrs.

\subsection{Decontextualization}
Recall from Section \ref{sec:models} that we produce \cwrs from bidirectional LMs with character CNNs and two-layer LSTMs.
We propose a method to remove the dependence on context $c$ for the two LSTM layers (the CNN layer is already context-independent by design).
During LM training, the hidden states of each layer $h_t$ are computed by the standard LSTM equations:
\begin{align*}
    i_t &= \sigma \left (W_i x_t + + U_i h_{t-1} + b_i \right)\\
    f_t &= \sigma \left (W_f x_t + U_f h_{t-1} + b_f \right)\\
    \tilde{c}_t &= \tanh \left (W_c x_t + U_c h_{t-1} + b_c \right )\\
    o_t &= \sigma \left (W_o x_t + U_o h_{t-1} + b_o \right )\\
    c_t &= f_t \odot c_{t-1} + i_t \odot \tilde{c}_t\\
    h_t &= o_t \odot \tanh \left ( c_t \right )
\end{align*}
We produce contextless vectors from pretrained LMs by removing recursion in the computation (i.e. setting $h_{t-1}$ and $c_{t-1}$ to 0):
\begin{align*}
    i_t &= \sigma \left (W_i x_t + b_i \right )\\
    f_t &= \sigma \left (W_f x_t + b_f \right )\\
    \tilde{c}_t &= \tanh \left (W_c x_t + b_c \right )\\
    o_t &= \sigma \left (W_o x_t + b_o \right )\\
    c_t &=  i_t \odot \tilde{c}_t\\
    h_t &= o_t \odot \tanh \left ( c_t \right )
\end{align*}
This method is fast to compute, as it does not require recurrent computation and only needs to see each word once.
This way, each word is associated with a set of exactly three vectors from the three layers.

\begin{table}
\small
\centering
\begin{tabular}{ l | ccccccc }
\hline
Representations &  UD& SRL& NER \\\hline
GloVe &83.78 & 80.01& 83.90\\
fastText &83.93 &80.27 &83.40\\
Decontextualization & 86.88&81.41&87.72  \\
ELMo & 88.71 & 82.12  &88.65 \\
\end{tabular}
\caption{Context independent vs.\ dependent performance in English. All embeddings are 512-dimensional and trained on the same English corpus of approximately 50M tokens for fair comparisons.
We also concatenate 128-dimensional character LSTM representations with the word vectors in every configuration to ensure all models have character input. UD scores are LAS, and SRL and NER are $F_1$.}
\label{tab:decontext_benefits}
\end{table}

\paragraph{Performance of decontextualized vectors}
We perform a brief experiment to find what information is successfully retained by the decontextualized vectors, by using them as inputs to three tasks (in a monolingual English setting, for simplicity).
For Universal Dependencies (UD) parsing, semantic role labeling (SRL), and named entity recognition (NER), we used the standard train/development/test splits from UD English EWT \cite{zeman-EtAl:2018:K18-2} and Ontonotes \cite{Pradhan2013TowardsRL}.
Following \citet{mulcaire_NAACL2019}, we use strong existing neural models for each task: \citet{dozatmanning2017} for UD parsing, \citet{He2017DeepSR} for SRL, and \citet{Peters2017SemisupervisedST} for NER.

Table \ref{tab:decontext_benefits} compares the decontextualized vectors with the original \cwrs (ELMo) and the conventional word type vectors, GloVe \cite{Pennington2014GloveGV} and fastText \cite{bojanowski2017enriching}.
In all three tasks, the decontextualized vectors substantially improve over fastText and GloVe vectors, and perform nearly on  par with contextual ELMo.
This suggests that while part of the advantage of \cwrs is in the incorporation of context, they also  benefit from rich context-\emph{independent} representations present in deeper networks.

\subsection{Word Translation Test}
\label{sec:wordtranslation}
Given the decontextualized vectors from each layer of the bidirectional language models, we can measure the crosslingual lexical correspondence in the multilingual \cwrs by performing word translation. 
Concretely, suppose that we have training and evaluation word translation pairs from the source to the target language.
Using the same word alignment objective discussed as in Section \ref{sec:multilingual_cwrs}, we find a linear transform by aligning the decontextualized vectors for the training source-target word pairs.
Then, we apply this linear transform to the decontextualized vector for each source word in the evaluation pairs. The closest target vector is found using the cross-domain similarity local scaling (CSLS) measure \cite{conneau2017word}, which is designed to remedy the hubness problem (where a few ``hub'' points are nearest neighbors to many other points each) in word translation by normalizing the cosine similarity according to the degree of hubness.

We again take the dictionaries from \citet{conneau2017word} with the given train/test split, and always use English as the target language.
For each language, we take all words that appear three times or more in our LM training data and compute decontextualized vectors for them. Word translation is evaluated by choosing the closest vector among the English decontextualized vectors.

\subsection{Results}
\begin{table}
\small
\centering
\begin{tabular}{ l | cccccc }
\hline
Vector &  \deu & \spa & \fra & \ita & \por & \swe \\\hline
fastText & 31.6&54.8&56.7&50.2&55.5	&43.9\\\hline
\multicolumn{7}{c}{ELMos} \\\hline
Layer 0 & 19.7& 41.5&41.1 & 36.9 & 44.6&27.5 \\
Layer 1 &24.4&46.4&47.6&44.2&48.3&36.3\\
Layer 2 &19.9&40.5&41.9&38.1&42.5&30.9\\ 
\multicolumn{7}{c}{Rosita}\\\hline
Layer 0 &37.9&\textbf{56.6}&\textbf{58.2}&57.5&\textbf{56.6}&50.6\\
Layer 1 &\textbf{40.3}&56.3&57.2&\textbf{58.1}&56.5&\textbf{53.7}\\
Layer 2 &38.8&51.1&52.7&53.6&50.7&50.8\\\hline
\end{tabular}
\caption{Crosslingual alignment results (precision at 1) from decontextual probe. Layers 0, 1, and 2 denote the character CNN, first LSTM, and second LSTM layers in the language models respectively. }
\label{tab:alignment}
\end{table}
We present word translation results from our decontextual probe in Table \ref{tab:alignment}.
We see that the first LSTM layer generally achieves the best crosslingual alignment both in ELMos and Rosita.
This finding mirrors recent studies on layerwise transferability; representations from the first LSTM layer in a language model are most transferable across a range of tasks \cite{Liu2019LinguisticKA}.
Our decontextual probe demonstrates that the first LSTM layer learns the most generalizable representations not only across tasks but also across languages.
In all six languages, Rosita (joint LM training approach)  outperforms ELMos (retrofitting approach) and the fastText vectors.
This shows that for the polyglot (jointly trained) LMs, there is a preexisting similarity between languages' vector spaces beyond what a linear transform provides. The resulting language-agnostic representations lead to polyglot training's success in low-resource dependency parsing.

\section{Further Related Work}
In addition to the work mentioned above, much previous work has proposed techniques to transfer knowledge from a high-resource to a low-resource language for dependency parsing.
Many of these methods use an essentially (either lexicalized or delexicalized) joint polyglot training setup (e.g., \citealp{mcdonald-petrov-hall:2011:EMNLP,cohen2011unsupervised,duong-EtAl:2015:ACL-IJCNLP,Guo2016ARL,vilares-etal-2016-one,falenska-cetinoglu-2017-lexicalized} as well as many of the CoNLL 2017/2018 shared task participants: \citet{lim-poibeau-2017-system,vania-etal-2017-uparse,de-lhoneux-etal-2017-raw,Che2018ElmoUD,wan-etal-2018-ibm,smith-etal-2018-82,lim-etal-2018-sex}).
Some use typological information to facilitate crosslingual transfer (e.g., \citealp{naseem-etal-2012-selective,tackstrom-etal-2013-target,zhang-barzilay-2015-hierarchical,galactic, rasooli-collins-2017-cross,Ammar2016ManyLO}).
Others use bitext \cite{zeman-EtAl:2018:K18-2}, manually-specified rules \cite{naseem-etal-2012-selective}, or surface statistics from gold universal part of speech \cite{surface,wang-eisner-2018-synthetic} to map the source to target.
The methods examined in this work to produce multilingual \cwrs do not rely on such external information about the languages, and instead use relatively abundant LM data to learn crosslinguality that abstracts away from typological divergence.

Recent work has developed several probing methods for (monolingual) contextual representations \cite{Liu2019LinguisticKA,structuralprobe,Tenney2019BERTRT}.
\citet{Wada2018UnsupervisedCW} showed that the (contextless) input and output word vectors in a polyglot word-based language model manifest a certain level of lexical correspondence between languages.
Our decontextual probe demonstrated that the \textit{internal} layers of polyglot language models capture crosslinguality and produce useful multilingual \cwrs for downstream low-resource dependency parsing.

\section{Conclusion}
We assessed recent approaches to multilingual contextual word representations, and compared them in the context of low-resource dependency parsing.
Our parsing results illustrate that a joint training approach for polyglot language models outperforms a retrofitting approach of aligning monolingual language models. Our decontextual probe showed that jointly trained LMs learn a better crosslingual lexical correspondence than the one produced by aligning monolingual language models or word type vectors. 
Our results provide a strong basis for multilingual representation learning and for further study of crosslingual transfer in a low-resource setting beyond dependency parsing.

\section*{Acknowledgments}
The authors thank Nikolaos Pappas and Tal Schuster as well as the anonymous reviewers for their helpful feedback.
This research was funded in part by NSF grant IIS-1562364, a Google research award to NAS, the Funai Overseas Scholarship to JK, and the NVIDIA Corporation through the donation of a GeForce GPU.

\bibliography{conll-2019}
\bibliographystyle{acl_natbib}

\clearpage
\appendix
\section{Training Parameters}
\label{appendix_a}

In this section, we provide hyperparameters used in our models and training details for ease of replication.

\subsection{Language Models}
\begin{table}
\small
\centering
\begin{tabular}{ |l p{0.3\linewidth}|}
\hline
\multicolumn{2}{|c|}{Character CNNs}\\
Char embedding size & 16\\
(\# Window Size, \# Filters) & (1, 32), (2, 32), (3, 68), (4, 128), (5, 256), 6, 512), (7, 1024)\\ 
Activation & Relu\\
\hline
\multicolumn{2}{|c|}{Word-level LSTM}\\
LSTM size & 2048\\
\# LSTM layers & 2\\
LSTM projection size & 256\\
Use skip connections & Yes\\
Inter-layer dropout rate& 0.1\\
\hline
\multicolumn{2}{|c|}{Training}\\
Batch size & 128\\
Unroll steps (Window Size) & 20\\
\# Negative samples & 64\\
\# Epochs & 10\\
Adagrad \cite{journals/jmlr/DuchiHS11} lrate& 0.2\\
Adagrad initial accumulator value& 1.0 \\
\hline
\end{tabular}
\caption{Language model hyperparameters.}
\label{lm-hyp}
\end{table}

Seen in Table \ref{lm-hyp} is a list of hyperparameters for our language models. 
We use the publicly available code of \citet{Peters2018} for training.\footnote{\url{github.com/allenai/bilm-tf}} 
Following \citet{mulcaire_NAACL2019}, we reduce the LSTM and projection sizes to expedite training and to compensate for the greatly reduced training data---the hyperparameters used in \citet{Peters2018} were tuned for the One Billion Word Corpus \cite{chelba2013one}, while we used only 5\% as much text (approximately 50M tokens) per language. Contextual representations from language models trained with even less text are still effective \cite{Che2018ElmoUD,Schuster2019CrossLingual}, suggesting that the method used in this work would apply to even lower-resource languages that have scarce text in addition to scarce or nonexistent annotation, though at the cost of some of the performance.

\subsection{UD Parsing}
\begin{table}
\small
\centering
\begin{tabular}{ |l l|}
\hline
\multicolumn{2}{|c|}{Input}\\
Input dropout rate & 0.3\\
\hline
\multicolumn{2}{|c|}{Word-level BiLSTM}\\
LSTM size & 400\\
\# LSTM layers & 3\\
Recurrent dropout rate & 0.3\\
Inter-layer dropout rate & 0.3\\
Use Highway Connection& Yes\\
\hline
\multicolumn{2}{|c|}{Multilayer Perceptron, Attention}\\
Arc MLP size & 500\\ 
Label MLP size & 100\\ 
\# MLP layers & 1\\
Activation & Relu\\
\hline
\multicolumn{2}{|c|}{Training}\\
Batch size & 80\\
\# Epochs & 80\\
Early stopping & 50\\
Adam \cite{Kingma2015} lrate& 0.001\\
Adam $\beta_1$& 0.9\\
Adam $\beta_2$& 0.999\\
\hline
\end{tabular}
\caption{UD parsing hyperparameters.}
\label{ud-hyp}
\end{table}
For UD parsing, we generally follow the hyperparameters used for the dependency parsing demo in AllenNLP \cite{Gardner2017AllenNLP}. See a list of hyperparameters in Table \ref{ud-hyp}.
We use stratified sampling so that each training mini-batch has an equal number of sentences from the source and target languages.

\subsection{Multilingual Word Vectors}
We train our word type representations used for non-contextual baselines with fastText \cite{bojanowski2017enriching}. We use window size 5 and a minimum count of 5, with 300 dimensions.

\section{Other Low-Resource Simulations}
\label{appendix_b}
In addition to the 100-sentence condition, we simulated low-resource experiments with 500 and 1000 sentences of target language data, and zero-target-treebank experiments in which the parser was trained with only source language data, but with multilingual representations allowing crosslingual transfer. See Table \ref{tab:simextra} for these results.
The additional low-resource results confirm our analysis in Section \ref{sec:sim_results}: polyglot training is more effective the less target-language data is available, with a slight advantage for related languages.

{
\renewcommand{\arraystretch}{1.2}
\renewcommand{\tabcolsep}{5pt}
\begin{table*}[!t]
    \small
    \begin{center}
    \begin{tabulary}{\textwidth}{|l|cc|ccc|ccc|ccc|}
    \hline
    &
    \multicolumn{2}{c|}{$|D_{\tau}| = 0$}   & \multicolumn{3}{c|}{$|D_{\tau}| = 100$} &\multicolumn{3}{c|}{$|D_{\tau}| = 500$} & \multicolumn{3}{c|}{$|D_{\tau}| = 1000$} \\ 
    target& +eng & +rel. & mono & +eng & +rel. & mono & +eng & +rel. & mono &+eng & +rel. \\ \hline
    \ara & 10.31 & \textbf{20.47} & 62.50&73.39&\textbf{73.43}&76.15 & \textbf{79.55} & 79.16 & 79.43 & 81.38 & \textbf{81.49} \\
    \heb & 23.76 & \textbf{24.89} &64.53&74.86&\textbf{75.69}&79.27 & 82.35 & \textbf{82.92} & 82.59 & 84.59 & \textbf{84.70} \\
      \rowcolor[gray]{0.9}\hrv & 48.69 & \textbf{67.67} &63.49&79.21& \textbf{82.00} & 80.80 & 84.92 & \textbf{85.89} & 84.14 & 86.27 & \textbf{86.66} \\
      \rowcolor[gray]{0.9}\rus & 38.69 & \textbf{73.24} & 59.51&75.63 &\textbf{79.29} & 77.38 & 83.16 & \textbf{84.60} & 82.90 & 85.68 & \textbf{86.99} \\
    \nld & 61.68 & \textbf{72.90} &57.12&74.90&\textbf{77.01} &75.19 & \textbf{82.42} & 81.33 & 81.41 & \textbf{84.93} & 83.23 \\
    \deu & 51.18 & \textbf{68.66} &60.26&72.52&\textbf{73.45}& 72.94 & \textbf{77.88} & 77.68 & 76.46 & \textbf{78.67} & 78.57 \\
      \rowcolor[gray]{0.9} \spa & 55.85 & \textbf{75.88} & 64.97 & 80.86 & \textbf{81.55} & 79.67 & \textbf{84.88} & 84.63 & 82.97 & 86.69 & \textbf{86.81} \\
      \rowcolor[gray]{0.9} \ita & 59.71 & \textbf{78.12} & 69.17 & \textbf{84.63} & 83.51 & 82.96 & \textbf{88.96} & 87.91 & 87.03 & \textbf{90.22} & 89.32 \\
    \cmn & \textbf{8.16} & 5.34 & 53.36 & \textbf{63.63} & 61.47 & 71.94 & 74.88 & \textbf{74.98} & 77.42 & \textbf{79.07} & 78.96 \\
    \jpn & 4.12 & \textbf{11.66} & 72.37 & \textbf{80.94} & 80.24 & 86.20 & \textbf{87.74} & \textbf{87.74} & 88.74 & 89.08 & \textbf{89.32} \\
    \hline
    \end{tabulary}
    \caption{LAS for UD parsing with additional simulated low-resource and zero-target-treebank settings.}
    \label{tab:simextra}
    \end{center}
\end{table*}
}

\section{UD Treebanks}
\label{appendix_d}

Additional statistics about the languages and treebanks used are given in Table \ref{tab:lang_list_supp}.
{
\renewcommand{\arraystretch}{1.2}
\begin{table*}[h]
    \small
    \begin{center}
    \begin{tabulary}{0.5\textwidth}{l|lll|l|lp{4cm}}
    \hline
    Lang & Code & WALS Genus & WALS 81A & Size (\# sents.)   & Treebank  & Genre\\ \hline\hline
    English & eng & Germanic & SVO &        &  EWT & blog, email, reviews, social \\ \hline
    \multicolumn{6}{c}{Simulation Pairs} \\ \hline
    Arabic & ara & Semitic & VSO/SVO &     & PADT & news \\
    Hebrew & heb & Semitic & SVO & \multirow{-2}{*}{5241} & HTB & news \\
%\rowcolor[gray]{0.9} Estonian & est & Finnic & SVO\\
%\rowcolor[gray]{0.9} Finnish & fin & Finnic &SVO \\
    \rowcolor[gray]{0.9} Croatian & hrv & Slavic & SVO &  & SET & news, web, wiki \\
    \rowcolor[gray]{0.9} Russian  & rus & Slavic & SVO & \multirow{-2}{*}{6983}  & SynTagRus&contemporary fiction, popular, science, newspaper, journal articles, online news\\
    Dutch   & nld & Germanic & SOV/SVO &  & Alpino & news \\
    German  & deu & Germanic & SOV/SVO & \multirow{-2}{*}{12269}  & GSD & news, reviews, wiki\\
    \rowcolor[gray]{0.9} Spanish & spa & Romance & SVO &   & GSD & blog, news, reviews, wiki \\
    \rowcolor[gray]{0.9} Italian & ita & Romance & SVO & \multirow{-2}{*}{12543}  &  ISDT & legal, news, wiki \\
    Chinese  & cmn & Chinese  & SVO &   & GSD & wiki \\
    Japanese & jpn & Japanese & SOV & \multirow{-2}{*}{3997}  & GSD & wiki \\ \hline
    \multicolumn{6}{c}{Truly Low Resource and Related Languages} \\ \hline
    \textbf{Hungarian}  & hun & Ugric   & SOV/SVO   & 910  & Szeged & news \\
    Finnish             & fin & Finnic  & SVO       & 12217 & TDT& news, wiki, blog, legal, fiction, grammar-examples\\ 
    \rowcolor[gray]{0.9}
    \textbf{Vietnamese} & vie & Viet-Muong  & SVO & 1400 & VTB & news\\
  %  \rowcolor[gray]{0.9}
    %Basque              & eus & Basque      & SOV & 5396 & BDT & news \\ \hline
    \textbf{Uyghur} & uig & Turkic  & SOV   &  1656 & UDT & fiction \\
    \textbf{Kazakh} & kaz & Turkic  & SOV (not in WALS)   & 31  & KTB& wiki, fiction, news \\
    Turkish         & tur & Turkic  & SOV       & 3685 & IMST&  nonfiction, news\\ 
    \end{tabulary}
    \caption{List of the languages and their UD treebanks used in our experiments. Each shaded/unshaded section corresponds to a pair of \textit{related} languages. WALS 81A denotes Feature 81A in WALS, Order of Subject, Object, and Verb \cite{wals}. Size represents the downsampled size in \# of sentences used for source treebanks.}
    \label{tab:lang_list_supp}
    \end{center}
\end{table*}
}

\section{Additional Experiments}
\label{appendix_e}
\paragraph{Semantic Role Labeling}
\begin{table}
\small
\centering
\begin{tabular}{ |l l|}
\hline
\multicolumn{2}{|c|}{Input}\\
Predicate indicator embedding size & 100\\
\hline
\multicolumn{2}{|c|}{Word-level Alternating BiLSTM}\\
LSTM size & 300\\
\# LSTM layers & 4\\
Recurrent dropout rate & 0.1\\
Use Highway Connection& Yes\\
\hline
\multicolumn{2}{|c|}{Training}\\
Batch size & 80\\
\# Epochs & 80\\
Early stopping & 20\\
Adadelta \cite{Zeiler2012ADADELTAAA} lrate& 0.1\\
Adadelta $\rho$ & 0.95\\
Gradient clipping & 1.0\\
\hline
\end{tabular}
\caption{SRL hyperparameters.}
\label{srl-hyp}
\end{table}
For SRL, we again follow the hyperparameters given in AllenNLP (Table \ref{srl-hyp}).
The one exception is that we used 4 layers of alternating BiLSTMs instead of 8 layers to expedite the training process. 

\paragraph{Named Entity Recognition}
\begin{table}
\small
\centering
\begin{tabular}{ |l l|}
\hline
\multicolumn{2}{|c|}{Char-level LSTM}\\
Char embedding size & 25\\
Input dropout rate & 0.5\\
LSTM size & 128\\
\# LSTM layers & 1\\
\hline
\multicolumn{2}{|c|}{Word-level BiLSTM}\\
LSTM size & 200\\
\# LSTM layers & 3\\
Inter-layer dropout rate & 0.5\\
Recurrent dropout rate & 0.5\\
Use Highway Connection & Yes\\
\hline
\multicolumn{2}{|c|}{Multilayer Perceptron}\\
MLP size & 400\\ 
Activation & tanh\\
\hline
\multicolumn{2}{|c|}{Training}\\
Batch size & 64\\
\# Epochs & 50\\
Early stopping & 25\\
Adam \cite{Kingma2015} lrate& 0.001\\
Adam $\beta_1$& 0.9\\
Adam $\beta_2$& 0.999\\
L2 regularization coefficient&0.001 \\
\hline
\end{tabular}
\caption{NER hyperparameters.}
\label{ner-hyp}
\end{table}
We again use the hyperparameter configurations provided in AllenNLP. See Table \ref{ner-hyp} for details.

\end{document}